\pdfoutput=1

\documentclass[11pt]{article}

\usepackage[]{EMNLP2023}

\usepackage{times}
\usepackage{latexsym}

\usepackage[T1]{fontenc}

\usepackage[utf8]{inputenc}

\usepackage{microtype}

\usepackage{inconsolata}

\usepackage{multirow}           
\usepackage{booktabs}           
\usepackage{arydshln}
\usepackage{graphicx}           
\usepackage{float}              
\usepackage{bm}                 
\usepackage[tikz]{bclogo}       
\usepackage{amsmath}            
\usepackage{amsfonts}           
\usepackage{pgfplots}           
\pgfplotsset{compat=1.18}
\usepackage{caption}
\usepackage{subcaption}         
\usepackage{CJK}
\usepackage{enumerate}

\usepackage[hang,flushmargin]{footmisc} 

\usepackage{algorithm} 
\usepackage{algpseudocode} 

\usepackage{bbding}

\definecolor{brickred}{HTML}{b92622}
\definecolor{midnightblue}{HTML}{005c7f}
\definecolor{salmon}{HTML}{f1958d}
\definecolor{burntorange}{HTML}{f19249}
\definecolor{junglegreen}{HTML}{4dae9d}
\definecolor{forestgreen}{HTML}{499c5e}
\definecolor{pinegreen}{HTML}{3d8a75}
\definecolor{seagreen}{HTML}{6bc1a2}
\definecolor{limegreen}{HTML}{97c65a}

\newcommand{\DatasetName}{Visual-C$^3$}

\title{Towards Real-World Writing Assistance: A Chinese Character Checking Benchmark with Faked and Misspelled Characters}

\author{
Yinghui Li$^{1}$\thanks{$^*$ indicates equal contribution.},
~Zishan Xu$^{1*}$, 
~Shaoshen Chen$^{1}$, 
~Haojing Huang$^{1}$,
~Yangning Li$^{1,5}$\\
~\textbf{Yong Jiang}$^{2}$,
~\textbf{Zhongli Li}$^{3}$,
~\textbf{Qingyu Zhou}$^{4,\dagger}$,
~\textbf{Hai-Tao Zheng}$^{1,5}$\thanks{ \; Corresponding authors: Hai-Tao Zheng and Qingyu Zhou\\qyzhgm@gmail.com, zheng.haitao@sz.tsinghua.edu.cn},
~\textbf{Ying Shen}$^{6}$\\
        $^{1}$Tsinghua Shenzhen International Graduate School, Tsinghua University \\ 
        $^{2}$Alibaba Group, $^{3}$Baidu Inc.,$^{4}$OPPO Research Institute, $^{5}$Peng Cheng Laboratory\\
        $^{6}$School of Intelligent Systems Engineering, Sun-Yat Sen University \\
        \tt{\{liyinghu20, xzs23\}@mails.tsinghua.edu.cn}
}

\begin{document}

\maketitle

\begin{CJK*}{UTF8}{gbsn}

\begin{abstract}
Writing assistance is an application closely related to human life and is also a fundamental Natural Language Processing (NLP) research field. 
Its aim is to improve the correctness and quality of input texts, with character checking being crucial in detecting and correcting wrong characters.
From the perspective of the real world where handwriting occupies the vast majority, characters that humans get wrong include faked characters (i.e., untrue characters created due to writing errors) and misspelled characters (i.e., true characters used incorrectly due to spelling errors). 
However, existing datasets and related studies only focus on misspelled characters mainly caused by phonological or visual confusion, thereby ignoring faked characters which are more common and difficult.
To break through this dilemma, we present \textbf{\DatasetName{}}, a human-annotated \textbf{Visual} \textbf{C}hinese \textbf{C}haracter \textbf{C}hecking dataset with faked and misspelled Chinese characters.  
To the best of our knowledge, \DatasetName{} is the first real-world visual and the largest human-crafted dataset for the Chinese character checking scenario. Additionally, we also propose and evaluate novel baseline methods on \DatasetName{}.
Extensive empirical results and analyses show that \DatasetName{} is high-quality yet challenging. The \DatasetName{} dataset and the baseline methods will be publicly available~\footnote{\url{https://github.com/THUKElab/Visual-C3}} to facilitate further research in the community.
\end{abstract}

\section{Introduction}\label{sec:intro}
With the continuous progress of social intelligence, texts on the Internet are growing explosively every day. Therefore, writing assistance that is to improve the correctness and quality of texts is becoming increasingly important~\cite{DBLP:journals/ce/StroblABDKPR19,DBLP:journals/corr/abs-2303-16726}, and has received more and more attention from researchers in recent years. Especially in the era of Large Language Models (LLMs), many studies have shown that writing assistance scenarios represented by text correction are particularly challenging for LLMs~\cite{DBLP:journals/corr/abs-2307-09007}. 
In the research field of writing assistance, the character checking task aims at detecting and correcting wrong characters in the given text and occupies a crucial position, as it ensures the correctness of the minimum atom (i.e., the characters) of texts~\cite{DBLP:journals/corr/abs-2204-03685}. Recently, for one of the most widely spoken languages in the world, large amounts of research have begun to focus on Chinese Character Checking, which is also well known as Chinese Spell Checking or Chinese Spelling Correction (CSC)~\cite{wu-etal-2013-integrating,DBLP:conf/acl-sighan/YuL14}. In this work, we also focus on the scene of Chinese Character Checking.

\begin{figure}[tbp!]
\centering
\includegraphics[width=\linewidth]{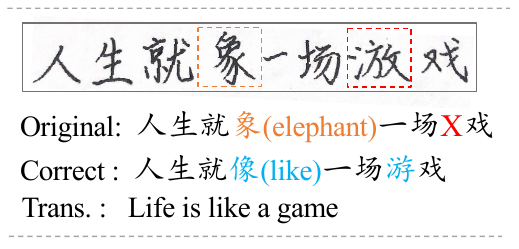}
\caption{Examples of Chinese faked characters (错字)  and misspelled characters (别字). \textcolor{orange}{Orange}/\textcolor{red}{red} represents the \textcolor{orange}{misspelled} character and the \textcolor{red}{faked} character.}

\label{fig:intro}
\end{figure}

Since Chinese Character Checking is a daily application closely related to human life, to promote its progress and development, we must consider the real-world application needs of humans for it. Therefore, a natural question arises: \textbf{What are the types of erroneous Chinese characters that humans would produce during the writing process?} Based on the observation of human writing habits, it is well known that there exist two main types of Chinese characters that humans get wrong in the real world, namely faked characters (错字) and misspelled characters (别字)~\cite{DBLP:journals/ijclclp/ChenB98}. As illustrated in Figure~\ref{fig:intro}, the misspelled character itself is a character that exists but is used incorrectly, the faked character is a non-existent character caused by incorrect writing (e.g., wrong use of radicals or wrong number of strokes). 
Authoritative Chinese linguistics studies~\cite{chinese_linguistics} have shown that faked characters appear more frequently than misspelled characters in the process of people's daily use of Chinese characters, and faked characters are often more difficult to detect than misspelled ones because faked characters are often caused by some very slight stroke errors.

As previously mentioned, the most widely studied task in the Chinese Character Checking scenario is the CSC task, which focuses on how to deal with misspelled characters caused by phonological or visual confusion~\cite{liu-etal-2010-visually,wang-etal-2019-confusionset}. Therefore, although faked characters are more common and challenging in the real world, researchers have not paid enough attention to how to handle the faked characters. The main reason for this dilemma is that the existing CSC data resources are all text-based. The biggest limitation of single text-modal data is that it can only represent characters that can be encoded by computers~\cite{DBLP:journals/patterns/LiuLTLZ22, DBLP:conf/icassp/LiCLXCZ23, DBLP:journals/corr/abs-2307-14878, DBLP:conf/semweb/ChenGFZCPLCZ23}, and faked characters themselves are not real and correct characters, and they obviously do not exist in the computer text encoding system. Hence, the traditional CSC datasets cannot represent faked characters, and the various CSC models proposed based on these text datasets cannot hold onto more complex and real scenarios. At this point, a pressing and significant problem is \textbf{how to expand and develop data resources for Chinese Character Checking to facilitate the automatic detection and correction of faked characters by models.}

Inspired by the enthusiasm to handle the faked characters, we propose to extend Chinese Character Checking to the visual modality, as images are the most direct and natural form to represent the faked characters. We construct a large-scale human-annotated \textbf{Visual} \textbf{C}hinese \textbf{C}haracter \textbf{C}hecking dataset, \textbf{\DatasetName{}}, which consists of 10,072 sentences represented by images and 12,019 wrong characters (including 5,670 misspelled and 6,349 faked characters) manually annotated by well-trained annotators. To the best of our knowledge, \DatasetName{} is the first real scene-oriented dataset that contains both faked and misspelled characters and also one of the largest human-annotated Chinese Character Checking datasets so far. Furthermore, to make future research on \DatasetName{} have more possibilities and potential, in addition to annotating sentence-level information for each image (this includes the golden sentence without error characters corresponding to the original content of the input image), we also annotate each picture at the character level. Specifically, referring to the data annotation methods in Computer Vision (CV), we segment each image into regions, so as to provide the accurate position and boundary information of each character on the image. Additionally, \DatasetName{} annotates wrong character type information (i.e., faked/misspelled) for more fine-grained analyses. Rich sentence/character-level annotation information makes our \DatasetName{} suitable for various NLP, CV, or multimodal models. 

Based on \DatasetName{}, we design the benchmark task to require the model to input an image containing sentences with wrong characters, and output the correct sentence without wrong characters corresponding to the input image in the form of text. Through this task, \DatasetName{} effectively assesses the detection and correction ability of Chinese Character Checking methods, especially for faked characters.
To verify the quality and challenge of \DatasetName{}, we design and implement two baseline methods with different paradigms and evaluate them on \DatasetName{}. Extensive experiments and detailed analyses demonstrate that \DatasetName{} is high-quality yet challenging. At the same time, the baselines also provide insightful and promising future directions. \textbf{Hopefully, we believe that the emergence of \DatasetName{} could promote the research of writing assistance to better adapt to the intelligence needed in the real world.}
\section{Related Works}
\subsection{Chinese Spell Checking}

Chinese Spell Checking or Chinese Spelling Correction (CSC) is a task to detect and correct erroneous characters in Chinese sentences~\cite{DBLP:conf/icassp/ZhangLZMLCZ23, DBLP:conf/emnlp/MaLSZHZLLLCZS22, DBLP:journals/corr/abs-2305-10819, DBLP:journals/corr/abs-2310-11671}, which plays an indispensable role in many downstream Natural Language Processing (NLP) applications~\cite{DBLP:journals/csur/DongLGCLSY23, DBLP:conf/sigir/LiLHYS022, DBLP:journals/tkde/LiHZZLLCZS23}. Although many advanced CSC methods~\cite{xu-etal-2021-read,huang-etal-2021-phmospell,li-etal-2022-learning-dictionary,li-etal-2022-past,DBLP:journals/corr/abs-2306-17447,DBLP:journals/corr/abs-2310-09119} have been proposed, building a high-quality CSC model still relies heavily on high-quality large-scale datasets. In recent years, several public CSC datasets have been proposed, which can be divided into two categories based on data content distribution: open-domain and specific-domain.

For open-domain, the most widely used are the SIGHAN datasets, which include SIGHAN13~\cite{wu2013chinese}, SIGHAN14~\cite{yu2014overview}, and SIGHAN15~\cite{tseng2015introduction}. 
In particular, SIGHAN datasets come from mistakes in essays written by teenage students (SIGHAN13) or Chinese as foreign language learners (SIGHAN14 and SIGHAN15). 
As for the specific-domain CSC datasets, MCSCSet~\cite{jiang2022mcscset} is a large-scale specialist-annotated dataset containing about 200K samples from a real-world medical application named Tencent Yidian. 
ECSpell~\cite{lv2023general} is a CSC dataset with three domains, law, medical, and official document. 
LEMON~\cite{wu2023rethinking} is a large-scale multi-domain dataset with natural spelling errors, which spans 7 domains, including game, encyclopedia, contract, medical, car, novel, and news.

However, the existing CSC datasets have one major limitation that cannot be ignored, that is, the modality of these datasets is limited to the single text modality.  The immediate dilemmas posed by this limitation are twofold.
First, the scope of error correction is limited. All existing CSC datasets do not cover text in images, i.e., they are unable to correct errors in realistic scenarios. The correction of errors in realistic scenarios is of paramount importance, while spelling errors in the real world do not only exist in text but also more widely in images, audio, and videos.
The second-fold dilemma is the incomplete type of error correction, i.e., the inability to handle the faked characters, whereas humans are more likely to make in daily life. The existing CSC datasets are all constructed in text form, so they cannot contain the faked characters at all. 
Therefore, to overcome the limitations described above, we construct \DatasetName{}, the first real-world visual and the largest human-crafted dataset for the Chinese Character Checking scenario.

\subsection{OCR Error Correction}

Optical Character Recognition (OCR) Error Correction refers to the process of detecting and rectifying errors in the text output produced by an OCR system. This task is somewhat related to Visual Chinese Character Checking, which our work focuses on. Therefore, it is necessary to introduce the related data resources of OCR error correction.
HANDS-VNOnDB3~\cite{nguyen2018database} has been presented to promote the studies on Vietnamese handwritten text recognition and to provide a benchmark to verify and compare the recognition performance of different approaches. It has handwritten images that contain 1,146 Vietnamese paragraphs of handwritten text comprising 7,296 lines, more than 480K strokes, and more than 380K characters. 
\citet{tanaka2022corpus} constructed a dataset based on the historical newspaper database Trove~\cite{cassidy2016publishing,sherratt2021glam} and public meeting articles in Australian historical newspapers~\cite{fujikawa1990public}, which contains 719 public meeting articles including 13,543 lines. 
Additionally, ICDAR2017~\cite{chiron2017icdar2017} comprises documents that were extracted from various sources, including a subset of the corpus related to the AmeliOCR project, as well as diverse collections like the French National Library and the British Library.

To the best of our knowledge, the existing OCR error correction datasets predominantly encompass language that is non-Chinese, noticeably lacking Chinese error correction data.
More importantly, existing OCR error correction data resources mainly focus on the challenges caused to the OCR system due to factors such as occlusion, shadow, and surrounding environment, while our \DatasetName{} focuses more on the same problem as the CSC task, i.e., wrong written Chinese characters caused by close similar pronunciation or shape.
\begin{figure*}[tbp!]
\centering
\includegraphics[width=\textwidth,height=\textheight,keepaspectratio]{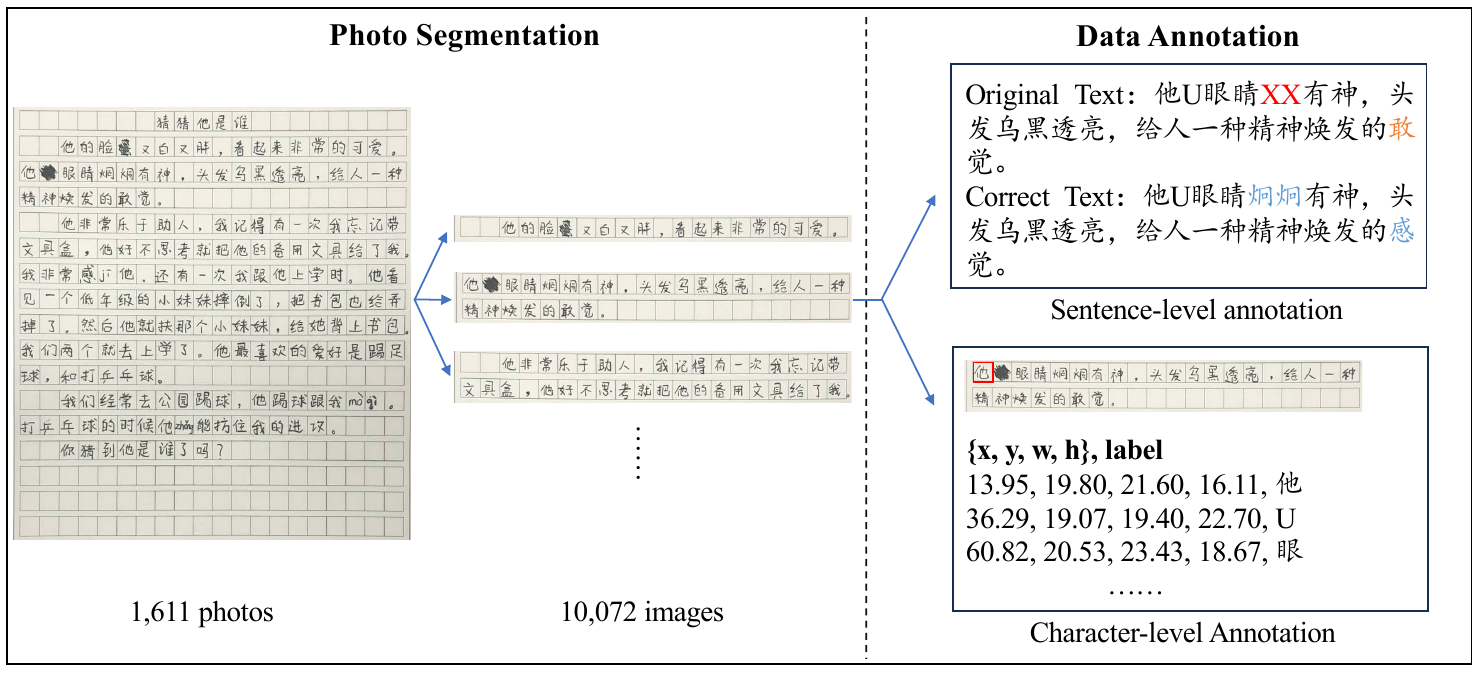}
\caption{Overview of the construction process of \DatasetName{}. ``U'' represents the unknown character, and ``X'' represents the faked character.}
\label{Figure:Annotation_Schema}
\end{figure*}

\section{The \DatasetName{} Dataset}
To facilitate read-world writing assistance, we propose \DatasetName{}, a Chinese Character Checking dataset composed of Sentences annotated with both faked characters and misspelled characters. 
\subsection{Dataset Construction}
\paragraph{Data Collection} Since Chinese Character Checking is a common daily application, when collecting raw data, we need to consider the daily life usage scenarios that really require it. This not only facilitates us to collect large amounts of data more conveniently, but also facilitates models developed based on our collected data to directly empower applications in daily life. Therefore, we focus on primary and secondary education, which is a real scenario that still requires students to do much writing and has high requirements on the accuracy of the characters written by students. Benefiting from the Chinese Character Checking models, in real primary and secondary school teaching, both teachers and students can detect and correct wrongly used characters more quickly and accurately, thereby achieving truly intelligent teaching.

Specifically, we cooperate with a Chinese language teaching and research group in a middle school and take photos of their students' handwritten essays as the raw data~\footnote{It is worth noting that we have signed a legal intellectual property agreement with the school and paid a data purchase fee of \$5 per essay to every student who provided valid data.}. There are two main reasons why we chose the photos of middle school students' handwritten essays as the raw data: (1) Photos of handwritten text are most consistent with real scenes and they can display faked and misspelled characters at the same time, while data in text format cannot represent faked characters. (2) The average Chinese character writing mastery level of middle school students determines that they will neither make simple mistakes that are too low-level nor make no mistakes at all, which ensures the challenge and usability of our data set. The entire data collection process lasted for 3 months, and we finally collected the photos of 5,692 handwritten essays from 389 students as our raw data.

\paragraph{Data Preprocessing} In order to ensure the quality of our dataset, we carefully check and filter the 5,692 original photos collected one by one. In particular, after observing the raw data, we identified three main categories of situations that we think may affect the dataset quality: (1) Students excessively daube and modify some characters during their writing process, seriously affecting the clarity of the photos and their characters. (2)  Some photos contain the teacher's red markings for faked and misspelled characters, which we believe will cause information leakage in the data sample. (3) Some photos are affected by many factors such as location and light during the shooting process, which affects the clarity of the photos and the completeness of the content of the essay. After our careful data cleaning, we finally retained 1,611 high-quality photos for the next step of annotation.

\paragraph{Annotation Schema}
To obtain sentence-level data, we segmented the 1,611 original photos into 10,072 images containing only one semantically complete sentence, as illustrated in Figure~\ref{Figure:Annotation_Schema}.
In the data annotation process, we utilize two annotation formats, namely sentence-level and character-level. 
For the sentence level, we annotate both the original text and the correct text. Note that the original text annotation is consistent with the content of the image, that is, the original text contains faked and misspelled characters. Particularly, for the faked characters, we marked them using the symbol ``X''. And for some characters that are difficult to recognize, we will directly mark them as ``Ut''.
At the character level, we annotate the position information for each character on the image. Specifically, we document the coordinate values (x, y) representing the top-left corner of each character, along with the length and width dimensions(w,h). This enables us to capture the precise positional information of each character, as depicted in Figure~\ref{Figure:Annotation_Schema}. Such character-level annotation is a common practice in the object detection field~\cite{DBLP:journals/ijcv/LongHY21}.

\paragraph{Annotation Workflow} As described in ``Annotation Schema'', there are two types of information we need to annotate, namely sentence-level text-based annotation for text and character-level image-based annotation. 
\begin{enumerate}[(1)]
    \item For the sentence-level annotation, we arranged 30 annotators and 10 senior annotation experts. All annotators are native Chinese speakers with sufficient linguistic foundation and are instructed in the principles and guidelines of annotation in detail. All senior annotation experts are Chinese teachers in our partner middle school. Specifically, each segmented image is independently annotated by three annotators and double-checked by one senior expert. The annotator is responsible for transcribing the content in the image into the original sentence containing faked characters represented by the symbol ``X'' and misspelled characters, and is responsible for modifying the original sentence into a correct sentence without wrong characters. Then, one annotator expert carefully checks the original/correct sentences for possible wrong or omissive annotations and makes the final decision in case three annotators have inconsistent correction results for the same image.
    \item For the character-level annotation, we employed 10 annotators and 2 senior experts professionally serving image segmentation. With the help of the specialized tool for localizing image areas, the annotation efficiency of annotators is greatly improved. Therefore, each image is annotated by an annotator using the tool to achieve the specific coordinate position information of each character on it, and then a senior expert checks the accuracy of the annotated coordinate information.
\end{enumerate}

To ensure the quality of the annotation workflow, we paid all annotators at market prices according to their workload (the number of images per hour) and we designed bonus rules for the senior annotators to check out annotation errors. In addition, we designed a batch annotation method, that is, we divided the entire raw data into 10 batches, and the data was annotated and submitted in batches. In the process of advancing the annotation workflow, we (i.e., the core authors)  will randomly select 20\% of the data submitted by senior annotation experts for sampling check. If the annotation accuracy of randomly selected samples is lower than 95\%, this batch will be returned for re-annotation. Overall, the entire annotation process of \DatasetName{} lasted about 3 months.

\subsection{Dataset Analysis}

\paragraph{Dataset Statistics}
\DatasetName{} consists of 10,072 sentences represented by images and 12,019 wrong characters are manually annotated by well-trained annotators. We randomly divided the training set, validation set, and test set according to the ratio of 3:1:1. We counted three attributes, namely average length, number of misspellings, and number of faked characters, respectively. As can be seen from Table~\ref{Table:Dataset_Statistics}, compared with the previously widely used CSC datasets, our \DatasetName{} is not only the first dataset containing faked characters, but its data size is also very competitive.

\paragraph{Dataset Quality}
Considering that the batch annotation method we designed has guaranteed the annotation accuracy to a certain extent, we further measure the agreements between multiple annotators. In particular, we calculate the Fleiss' kappa~\cite{moons2023measuring} to verify the annotator agreement of labeling the original/correct sentences of images, the result is 85.20\%, which indicates that our annotation can be regarded as ``almost perfect agreement''~\cite{landis1977measurement}.

\begin{table}[]
\centering
\resizebox{\columnwidth}{!}{%
\begin{tabular}{@{}ccccc@{}}
\toprule
Dataset            & \#Sent & Avg.Length & \#Misspelled & \#Faked \\ \midrule
SIGHAN2013         & 1,700   & 60.9       & 1,567         & -       \\
SIGHAN2014         & 4,499   & 49.7       & 5,893         & -       \\
SIGHAN2015         & 3,439   & 31.1       & 3,740         & -       \\
\textbf{\DatasetName{}} & 10,072  & 40.4       & 5,670         & 6,349    \\ \bottomrule
\end{tabular}
}

\caption{Statistics of CSC datasets. Column Sentence represents the number of samples in this dataset.}
\label{Table:Dataset_Statistics}
\end{table}

\begin{figure*}[h]
\centering
\includegraphics[width=\textwidth,height=\textheight,keepaspectratio]{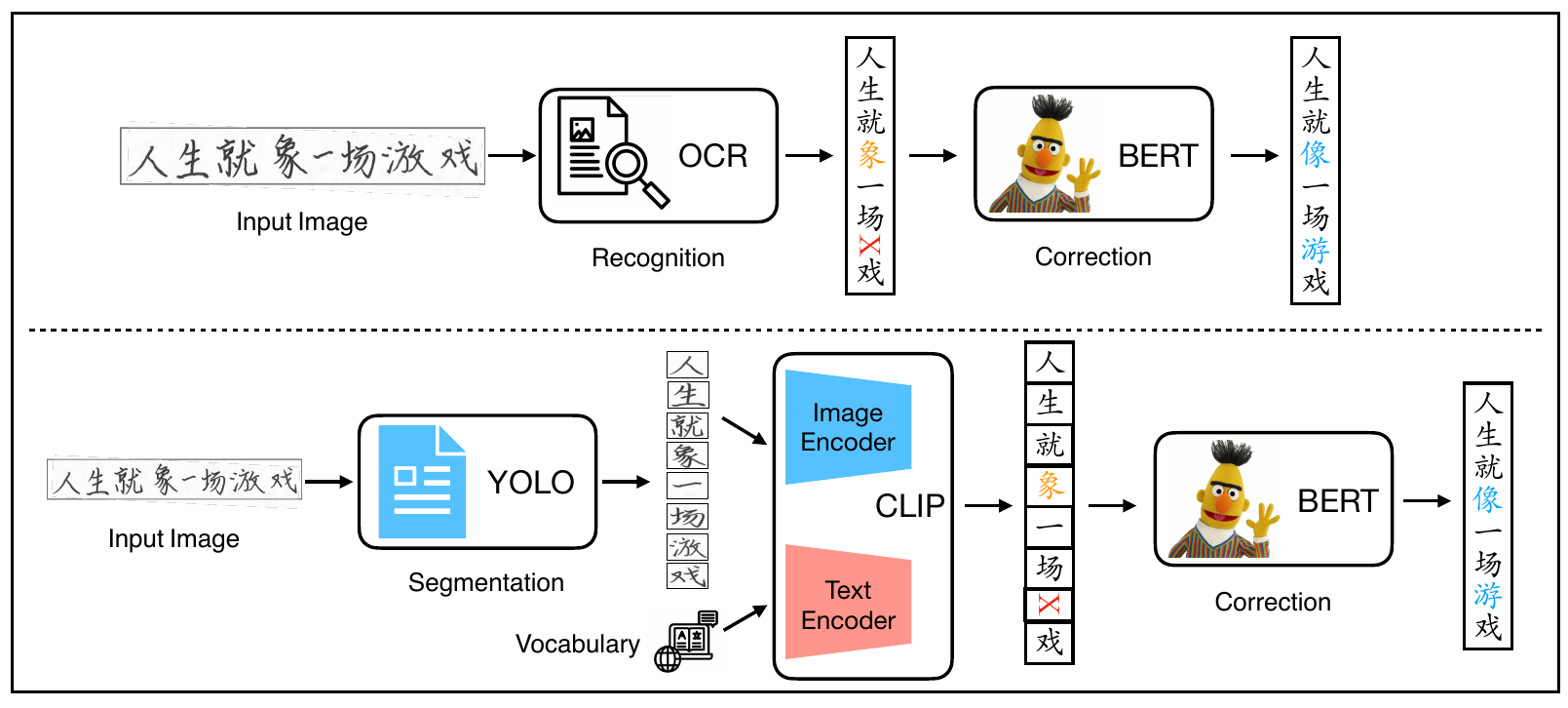}
\caption{Illustration of our designed baseline methods, namely OCR-based method (top) and CLIP-based method (bottom).}
\label{Figure:Method}
\end{figure*}

\subsection{Benchmark Settings}

\paragraph{Task Formulation} Our \DatasetName{} benchmark focuses on Visual Chinese Character Checking. To fully exploit \DatasetName{} so that it more comprehensively evaluates the model's Chinese Character Checking capabilities, especially the processing capabilities of the faked characters, we divide Visual Chinese Character Checking into two subtasks based on \DatasetName{}.
\begin{enumerate} [(1)]
    \item \textbf{Detection Subtask}: The inputs are images from \DatasetName{}, and the ideal outputs are corresponding text marked with faked and misspelled character positions. Note that the core of the detection subtask is to accurately identify which characters in the image are faked and which characters are misspelled. It does not require the model to know the correct characters corresponding to the faked or misspelled characters.
    \item \textbf{Correction Subtask}: After the detection subtask has accurately located which characters in the image are faked or misspelled, the correction subtask further requires the model to output a text with no wrong characters at all, that is, to correct the detected faked and misspelled characters.
\end{enumerate}

\paragraph{Evaluation Metrics} 
In terms of evaluation granularity, there are two levels of metrics, i.e., character and sentence levels. 
The sentence-level metric requires that all the wrong characters in a sentence are successfully detected and corrected.
And the sentence-level metric is more difficult than the character-level metric, because a sentence may have multiple wrong characters, and the character-level metric is more microscopic and detailed than the sentence-level metric.
Specifically, we calculate the Precision, Recall, and F1 score for the detection and correction subtasks. At the same time, in order to evaluate the model's processing capabilities for different types of characters in more detail, we also calculate metrics separately for faked and misspelled characters.

\section{Models and Baselines}

To reflect the usability of \DatasetName{} and provide reference ideas for future research on \DatasetName{}, we design and propose two baseline methods, namely OCR-based and CLIP-based methods, as illustrated in Figure~\ref{Figure:Method}.

\subsection{OCR-based Method}
The OCR-based method consists of two modules, namely the recognition module and the correction module. 
The recognition module is responsible for obtaining text content by identifying the characters in the input images, while the correction module corrects based on the output of the recognition module and outputs text without incorrect characters.

\paragraph{Recognition Module}
To recognize the Chinese characters on the images, we utilize an OCR model which has the ability to convert images into texts. Specifically, the input of this module is the image $I$ with $n$ characters and the output is the text $X = (x_{1} , x_{2}, \dots, x_{n})$ with faked and unknown characters. Consistent with the dataset annotation, the faked character is marked as ``X'' and the unknown character is marked as ``U''.

In particular, for traditional OCR methods, a great challenge with \DatasetName{} is how to recognize the faked characters.
Therefore, we propose two strategies to solve this dilemma. 
Heuristically, we treat characters that can not be well recognized by the OCR model as faked characters, because their failure to be recognized indicates that they are ambiguous. Specifically, any characters whose recognition module output confidence score is below a reasonable threshold $thr$ will be classified as faked characters.
Additionally, because the traditional OCR model lacks the inherent ability to recognize the faked characters, we employ our training datasets with customized vocabulary to train the OCR model. After fine-tuning, the OCR model in the recognition module will have the ability to recognize faked characters without any artificially set heuristic thresholds.




\paragraph{Correction Module}
The correction module is a sequential multi-class
labeling model based on transformers represented by BERT~\cite{devlin-etal-2019-bert,DBLP:journals/taslp/CuiCLQY21}. 
The input is the texts with faked and misspelled characters $X = (x_{1} , x_{2}, \dots, x_{n})$ and the output is a sequence of characters $Y = (y_{1} , y_{2}, \dots, y_{n})$.

For each character of the sequence, the probability of error correction is defined as
\begin{equation}
P (y_{i} = j | X) = \operatorname{softmax}(Wh_{i} + b)[j]
\end{equation}
where $ P_{c} (y_{i} = j | X)$ is the conditional probability that character $x_{i}$ is corrected as the character $j$ in the vocabulary, $ h_{i}$ denotes the hidden state, $ W$ and $ b$ are learnable parameters. Through this paradigm, the character that has the highest probability is selected from the vocabulary as output for character $x_{i}$. It is worth noting that the vocabulary of the correction module will be extended with the special tokens ``U'' and ``X'' to facilitate it to receive the output of the recognition module.

\subsection{CLIP-based Method}
The CLIP-based method is divided into three modules, which are the segmentation module, retrieval module, and correction module. The segmentation module is to segment a sentence-level image into character-level images. The retrieval module is to map each character-level image to its original Chinese character. And the function of the correction module is the same as the OCR-based method's.

\paragraph{Segmentation Module}
In the segmentation module, our primary objective is to accurately identify and arrange each character present in the image, following a traditional left-to-right and top-down ordering scheme.

To accomplish character segmentation, we employ an object detection approach capable of identifying all characters within the image.  This method enables us to extract the individual characters present in the image. Specifically, given the image $I$, we can obtain the coordinates of the upper left corner $(\bm{L}_X, \bm{L}_Y)$, as well as the width $\bm{W}$ and height $\bm{H}$ of each of the $n$ character-level sub-images segmented.

While the object detection model effectively identifies the characters in the image, arranging them in the correct order poses an enormous challenge.  Consequently, we propose a regularization sorting algorithm as outlined in Algorithm $\ref{algorithm}$ to establish the appropriate sequence of characters. Finally, the segmentation module will get a sequence of character-level images sorted according to the order of characters in the sentence.

\begin{algorithm} 
	\caption{Regularization Sorting} 
	 \label{algorithm}       
	\begin{algorithmic}[1] 
	\Require $\bm{L}_X$, $\bm{L}_Y$, $\bm{W}$, $\bm{H}$   
    \Ensure Sorted $\tilde{\bm{L}_X}$, $\tilde{\bm{L}_Y}$, $\tilde{\bm{W}}$, $\tilde{\bm{H}}$   
    \State $\tilde{\bm{L}_X}$ $\gets$ $\emptyset$, $\tilde{\bm{L}_Y}$ $\gets$ $\emptyset$, $\tilde{\bm{W}}$ $\gets$ $\emptyset$, $\tilde{\bm{H}}$ $\gets$ $\emptyset$    
    \Repeat
    \State Calculate the average value $\bar{\bm{M}}$ within the range of $\alpha$ for the minimum values of $\bm{L}_Y$
    \State Treat the index $\bm{i}$ of characters that are within a distance of $\beta$ from the mean $\bar{\bm{M}}$ 
    \State Sort $\bm{i}$ according to horizontal coordinate from small to large, it is put into $\tilde{X}$
    \State Take sorted coordinates according to $\bm{i}$ into $\tilde{\bm{L}_X}$, $\tilde{\bm{L}_Y}$, $\tilde{\bm{W}}$, $\tilde{\bm{H}}$ 
   \State Remove the coordinates already taken from $\bm{L}_X$, $\bm{L}_Y$, $\bm{W}$, $\bm{H}$
   \Until {$|\bm{L}_X|$ $\leq$ 0}
   \State\Return $\tilde{\bm{L}_X}$, $\tilde{\bm{L}_Y}$, $\tilde{\bm{W}}$, $\tilde{\bm{H}}$ 
   \end{algorithmic} 
\end{algorithm} 

\begin{table*}[h]
\centering
\resizebox{2\columnwidth}{!}{%
\begin{tabular}{@{}lcccccccccccc@{}}
\toprule
\multirow{3}{*}{Methods} & \multicolumn{3}{c}{Detection} & \multicolumn{3}{c}{Correction} & \multicolumn{3}{c}{Detection} & \multicolumn{3}{c}{Correction}\\
\multirow{3}{*}{} & \multicolumn{3}{c}{(Character-Level)} & \multicolumn{3}{c}{(Character-Level)} & \multicolumn{3}{c}{(Sentence-Level)} & \multicolumn{3}{c}{(Sentence-Level)}\\ 
\cmidrule(l){2-4} \cmidrule(l){5-7} \cmidrule(l){8-10} \cmidrule(l){11-13} 
    &  \hspace{0.2em}Prec.         & Rec.        & \hspace{0.2em}F1.
     & \hspace{0.2em}Prec.         & Rec.         & \hspace{0.2em}F1.
     & \hspace{0.2em}Prec.         & Rec.         & \hspace{0.2em}F1.
     & \hspace{0.2em}Prec.         & Rec.         & \hspace{0.2em}F1.          \\ \midrule
~~OCR-Based Method  &\hspace{0.5em}3.6           & 42.2        & \hspace{0.5em}6.6
            &\hspace{0.5em}2.0           & 23.5        & \hspace{0.5em}3.7
            &\hspace{0.5em}0.8           & \hspace{0.5em}2.8          & \hspace{0.5em}1.3
            &\hspace{0.5em}0.3           & \hspace{0.5em}0.9          & \hspace{0.5em}0.4          \\ 
\quad + Fine-tuned Recognition  & 16.0            & 56.3        & 25.0
                                                 & 14.1            & 49.3        & 21.9
                                                 & 11.6            & 23.4        & 15.5
                                                 & \hspace{0.5em}9.4             & 19.0           & 12.6         \\
\quad + Fine-tuned Recognition/Correction  & 16.2            & 55.8        & 25.1           & 14.2          
                                                 & 49.3            & 22.0        & 12.4           & 24.7
                                                 & 16.6            & 10.0        & 20.0           & 13.4 \\ \midrule 
~~CLIP-Based Method&\hspace{0.5em}9.8           & 55.7        & 16.8
                                                  &\hspace{0.5em}8.5           & 48.3        & 14.5
                                                  &\hspace{0.5em}5.4           & 13.5        &\hspace{0.5em}7.7      
                                                  &\hspace{0.5em}4.3           & 10.6        &\hspace{0.5em}6.0    \\
\quad + Fine-tuned Correction& 10.1                       & 56.9        & 17.2        
                                                  & \hspace{0.5em}8.7          & 48.9        & 14.8        
                                                  & \hspace{0.5em}5.5          & 13.5        & \hspace{0.5em}7.8
                                                  & \hspace{0.5em}4.7          & 11.5        & \hspace{0.5em}6.7   \\       \bottomrule
\end{tabular}
}
\caption{Performance of different methods on the misspelled characters of \DatasetName{} test set.}
\label{Table: main_miss}
\end{table*}


\begin{table*}[h]
\centering
\resizebox{2\columnwidth}{!}{%
\begin{tabular}{@{}lcccccccccccc@{}}
\toprule
\multirow{3}{*}{Methods} & \multicolumn{3}{c}{Detection} & \multicolumn{3}{c}{Correction} & \multicolumn{3}{c}{Detection} & \multicolumn{3}{c}{Correction}  \\
 & \multicolumn{3}{c}{(Character-Level)} & \multicolumn{3}{c}{(Character-Level)} & \multicolumn{3}{c}{(Sentence-Level)} & \multicolumn{3}{c}{(Sentence-Level)} \\ 
 \cmidrule(l){2-4} \cmidrule(l){5-7} \cmidrule(l){8-10} \cmidrule(l){11-13} 
                            & \hspace{0.2em}Prec.         & Rec.        & \hspace{0.2em}F1.
                             & \hspace{0.2em}Prec.         & Rec.        & \hspace{0.2em}F1.
                             & \hspace{0.2em}Prec.         & Rec.        & \hspace{0.2em}F1.
                             & \hspace{0.2em}Prec.         & Rec.        & \hspace{0.2em}F1.          \\ \midrule
~~OCR-Based Method      & \hspace{0.5em}3.6           & 36.0          & \hspace{0.5em}6.5
                & \hspace{0.5em}0.3           & \hspace{0.5em}2.9          & \hspace{0.5em}0.5
                & \hspace{0.5em}0.3           & \hspace{0.5em}0.8          & \hspace{0.5em}0.4
                & \hspace{0.5em}0.0           & \hspace{0.5em}0.0          & \hspace{0.5em}0.0            \\ \midrule
\quad + Fine-tuned Recognition        & 13.1          & 20.9        & 16.1
                                                     & \hspace{0.5em}5.9           & \hspace{0.5em}9.3          & \hspace{0.5em}7.2         & \hspace{0.5em}8.6           & \hspace{0.5em}9.1          & \hspace{0.5em}8.8         & \hspace{0.5em}4.7           & \hspace{0.5em}5.0            & \hspace{0.5em}4.9   \\
 \quad + Fine-tuned Recognition/Correction               & 13.1          & 20.9        & 16.1        
                                                     & \hspace{0.5em}7.1             & 11.4         & \hspace{0.5em}8.8
                                                     & \hspace{0.5em}8.6           & \hspace{0.5em}9.1       & \hspace{0.5em}8.8
                                                     & \hspace{0.5em}6.1           & \hspace{0.5em}6.4       & \hspace{0.5em}6.2\\  \midrule
~~CLIP-Based Method        & 14.3          & 15.5        & 14.9        
                                                      & \hspace{0.5em}6.7           & \hspace{0.5em}7.2          & \hspace{0.5em}6.9         & \hspace{0.5em}7.6           & \hspace{0.5em}8.5          & \hspace{0.5em}8.0         & \hspace{0.5em}4.4           & \hspace{0.5em}4.9          & \hspace{0.5em}4.6   \\
\quad + Fine-tuned Correction               & 14.3          & 15.5        & 14.9        & \hspace{0.5em}9.2           
                &\hspace{0.5em}9.9          & \hspace{0.5em}9.5         & \hspace{0.5em}7.6           & \hspace{0.5em}8.5          
                &\hspace{0.5em}8.0           & \hspace{0.5em}5.8           & \hspace{0.5em}6.4          & \hspace{0.5em}6.1 \\ \bottomrule
\end{tabular}
}
\caption{Performance of different methods on the faked characters of \DatasetName{} test set.}
\label{Table: main_fake}
\end{table*}

\paragraph{Retrieval Module}
After obtaining the images of each character sequentially, we carry out the image-text retrieval task based on CLIP~\cite{pmlr-v139-radford21a}.
The CLIP model usually has a text encoder and an image encoder to obtain the respective representations of texts and images, and then we can match or retrieve texts based on images according to the similarity between the representations of images and texts, this is the core objective of our designed image-text retrieval task.
Particularly, we train the CLIP model from scratch on \DatasetName{}, giving it the ability to retrieve corresponding Chinese characters based on images, especially the ability to identify faked characters. Through the retrieval module, we obtain the text including misspelled and fake characters after this module.

For the training of our used CLIP model, we instruct the text encoder to align itself with the image embedding by maximizing the cosine similarity between paired image/text embeddings within the batch, while simultaneously minimizing the cosine similarity of unpaired image/text within the batch. We optimize the CLIP model with the similarity score utilizing the contrastive loss:
\begin{equation}
L = -\frac{1}{n}\sum_{j=1}^{n}\log_{}{\frac{exp(sim(z_{j}^{t},z_{i}^{t})/\tau )}{ {\textstyle \sum_{k=1}^{n}exp(sim(z_{j}^{t},z_{i}^{k})/\tau) } } }    
\end{equation}
where $z^{t} = [z_{1}^{t}, z_{2}^{t}, \dots,z_{n}^{t}] $ represents the latent representations of texts, while $z^{i} = [z_{1}^{i}, z_{2}^{i}, \dots,z_{n}^{i}] $ represents those of images within a mini-batch comprising n samples.

\paragraph{Correction Module}
The function and implementation of this part module are the same as the correction module of the OCR-based method.

\section{Experiments and Analyses}
In this section, we introduce the details of our experiments to evaluate our proposed baseline methods.

\subsection{Implementation Details}
All the models presented in this paper are implemented using Python (Version 3.7.15) and the PyTorch framework (Version 1.12.1). 
For the OCR-based method, we select the PaddleOCRv3 of handwriting~\cite{DBLP:journals/corr/abs-2206-03001} to be the recognition module. 
If the recognition module is not fine-tuned, the faked characters will be classified by the $thr$ of 0.2.
We utilize the advanced and widely used YOLOv8 model~\footnote{\url{https://github.com/ultralytics/ultralytics}} to segment sentence-level images into character-level images.
For the implementation of our CLIP model in the retrieval module, we initialize the image encoder and text encoder with the ResNet-50~\cite{DBLP:conf/cvpr/HeZRS16} and RoBERTa-base~\cite{DBLP:journals/corr/abs-1907-11692}.
As for the correction module, we utilize the $BERT_{BASE}$~\cite{devlin-etal-2019-bert} which has 12 transformer layers with 12 attention heads.

Regarding the fine-tuning details, the recognition module of the OCR-based baseline is trained over 500 epochs, with a learning rate of 4e-5 and a batch size of 50. For the CLIP-based baseline, the detection module is trained for 2,000 epochs, employing a learning rate of 5e-5 and a batch size of 256. Additionally, the correction module is fine-tuned for 10 epochs, using a learning rate of 5e-5 and a batch size of 4.

\subsection{Main Results}
From Table~\ref{Table: main_miss} an Table~\ref{Table: main_fake}, we have the following observations:
\begin{enumerate}
    \item When not fine-tuned, the pre-trained OCR model performs extremely poorly on \DatasetName{}, which indicates that existing OCR methods cannot work well on our dataset and reflects the challenging nature of our dataset. In particular, for the faked characters, the performance of our proposed baselines is still unsatisfactory even after fine-tuning. Therefore,  how to empower machines to process A characters our work focuses on is obviously very urgent and important for the community.
    \item For the misspelled characters, we find that the models' recall is much higher than its precision. This is because the model corrects a large number of characters, thereby incorrectly modifying many correct characters. Therefore, the poor performance of the BERT-based correction module on the misspelled characters indicates that the text content of \DatasetName{} is also very difficult for mainstream CSC models, which once again reflects the challenge of our dataset.
    \item We are pleasantly surprised to find that the performance of the CLIP-based method we proposed is not very poor, which shows that our innovative idea of identifying the faked characters through image and text retrieval is completely feasible, which is also instructive for future research on the \DatasetName{} dataset.
\end{enumerate}

\subsection{Performance Analysis}
\begin{table}[h]
\centering
\resizebox{0.48\textwidth}{!}{
\begin{tabular}{@{}lcccc@{}}
\toprule
Methods & Misspelled & Faked &Correct & Average   \\
 \midrule
Fine-tuned OCR     & 0.694          & 0.209         & 0.944
             & 0.929               \\ 
Fine-tuned CLIP    & 0.732          & 0.155         & 0.929  & 0.914              \\ 
\bottomrule
\end{tabular}}
\caption{The character accuracy of the OCR model and the CLIP model. The total numbers of misspelled, faked, and correct characters in the test set of \DatasetName{} are 788, 1,223, and 79,141 respectively.}
\label{Table: OCR_CLIP_Performance}
\end{table}

\paragraph{The OCR and CLIP Performance: } Table~\ref{Table: OCR_CLIP_Performance} reports the performance of our fine-tuned OCR model and CLIP model, i.e., their character recognition (or retrieval) accuracy. After fine-tuning on \DatasetName{}, both the OCR model and the CLIP model have considerable average character recognition accuracy and a certain ability to distinguish the faked characters. Of course, we have to admit that compared with the misspelled and correct characters, our fine-tuned models' processing ability for the faked characters is still much inferior. Obviously, simple fine-tuning is not enough. We encourage subsequent researchers to make greater innovations in the model structure to obtain better performance of the faked characters on the \DatasetName{} dataset.

\begin{table}[h]
\centering
\setlength{\tabcolsep}{1mm}{
\begin{tabular}{@{}lccccccccc@{}}
\toprule
\multirow{2}{*}{Character} & \multicolumn{3}{c}{Correction} & \multicolumn{3}{c}{Correction}   \\
 \multirow{2}{*}{Type}& \multicolumn{3}{c}{(Character-Level)} & \multicolumn{3}{c}{(Sentence-Level)}  \\ \cmidrule(l){2-4} \cmidrule(l){5-7}
& \hspace{0.2em}Prec.          & Rec.         & \hspace{0.2em}F1.         & 
  \hspace{0.2em}Prec.          & Rec.         & \hspace{0.2em}F1.\\ \midrule
Misspelled     & 72.7          & 47.3         & 57.3
             & 52.7          & 40.4         & 45.8          \\ 
Faked    & 63.8          & 63.8         & 63.8
             & 58.4          & 58.4         & 58.4         \\ 
\bottomrule
\end{tabular}}
\caption{The performance upper bounds of different methods for different types of characters.}
\label{Table: Corr_Upper_Bound}
\end{table}
  
\paragraph{Correction Upper Bound: } To further measure the difficulty of the text content of our dataset for existing error correction methods, we study the performance upper bound of the correction module in our baselines. Specifically, we directly input the original text annotated in \DatasetName{} into the correction module to observe its correction performance. Note that we only report the performance of the fine-tuned correction module. From Table~\ref{Table: Corr_Upper_Bound}, we know that BERT's performance on \DatasetName{} is not very high, and BERT can achieve at least a score of 63.4 or more on sentence-level correction F1 on previously widely used CSC datasets such as SIGHAN13/14/15~\cite{wu2013chinese,yu2014overview,tseng2015introduction}. This performance gap indicates that the text content of our dataset is more complex than the previous CSC datasets and we think this difficulty stems from the fact that our dataset is collected from completely real scenes.

\begin{figure}[h]
\centering
\includegraphics[width=0.5\textwidth,height=\textheight,keepaspectratio]{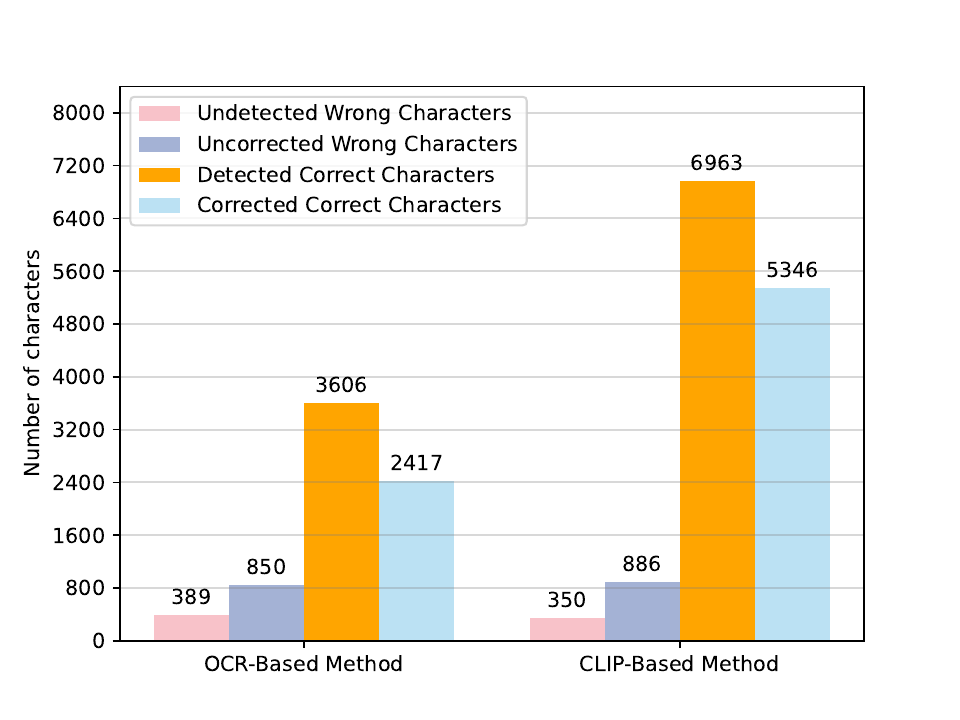}
\caption{Statistics of characters that are mishandled. The total numbers of wrong and correct characters in the test set of \DatasetName{} are 2,011, and 79,141 respectively.}
\label{Figure:Error_Analysis}
\end{figure}

\begin{figure*}
    \centering
    \includegraphics[width=1.98\columnwidth]{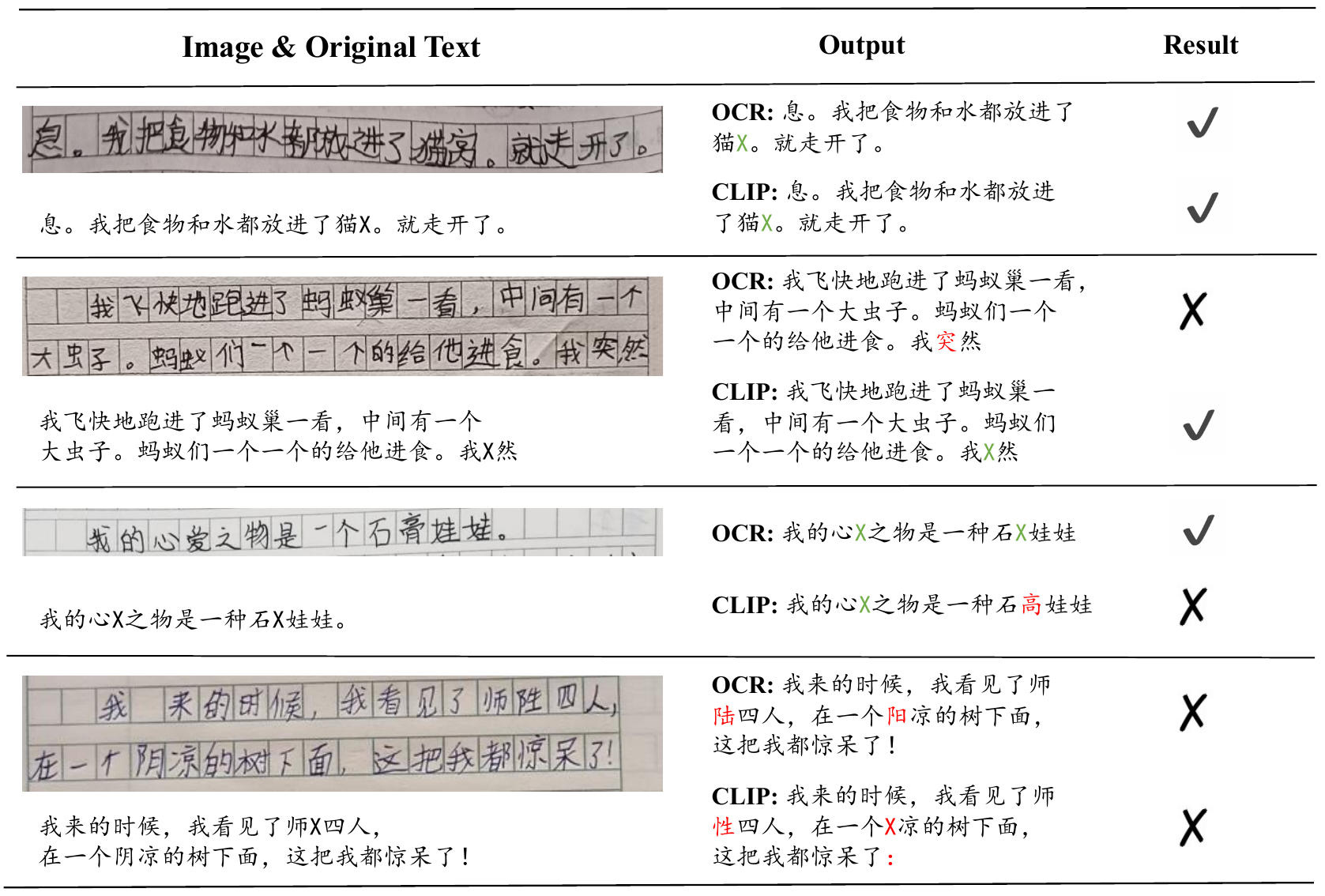}
    \caption{
    Some examples of our designed baselines. A \Checkmark mark indicates that the output of the corresponding method is correct, and a \XSolidBrush mark means that the output of the corresponding method is problematic.}
    \label{fig:case_study}
\end{figure*}

\begin{figure*}[h]
\centering
\includegraphics[width=1\textwidth,height=\textheight,keepaspectratio]{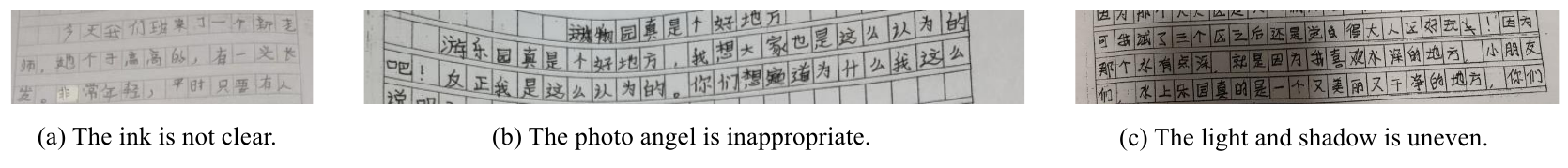}
\caption{Representatives of hard samples from the \DatasetName{} dataset.}
\label{fig:dataset_challenge}
\end{figure*}

\subsection{Error Analysis}
In order to conduct fine-grained analyses of what our proposed baseline methods do not do well, we perform error analysis on the fine-tuned OCR and CLIP-based methods. As shown in Figure~\ref{Figure:Error_Analysis}, we count the cases where different methods mishandle wrong characters (including misspelled and faked characters) and correct characters. We notice that whether it is the OCR-based or CLIP-baed method, they tend to detect or correct a large number of correct characters wrongly (it should be emphasized that the proportion of correct characters that are mishandled is not high). Based on our observations, we think that this kind of error mainly comes from the recognition module of the OCR-based method or the segmentation and retrieval modules of the CLIP-based method. Especially for the CLIP-based method, although we innovatively propose the method of image and text retrieval to identify the faked characters in images, the pipeline paradigm of first segmenting the sentence-level image into small character-level images and then retrieving will result in a certain degree of error accumulation. For the OCR-based method, the accuracy of the recognition module also determines the performance ceiling of the entire method to a certain extent. Therefore, studying how to identify the faked characters from images in an end-to-end manner will be a challenging and valuable research direction for the \DatasetName{} dataset in the future.

\subsection{Case Study}
\paragraph{Model Cases: }In Figure~\ref{fig:case_study}, we select some examples of successful and failed OCR/CLIP model processing for demonstration. From these cases, we know that after fine-tuning on the \DatasetName{} dataset, both the traditional OCR model and our newly proposed CLIP-based image-text retrieval model have the ability to recognize the faked characters in images. For future studies on \DatasetName{}, we think there are two ideas that can improve model performance. First, how can we make the model better handle complex characters with many strokes, such as ``突'' in the second case and ``膏'' in the third case? Second, it is crucial to improve the model to distinguish between the faked characters and misspelled characters with similar strokes. For example, in the fourth case, the model should detect the character in the image as a faked character, but it instead gives a ``陆'' with similar strokes as the output, which would lead to a decrease in the model's faked character detection performance.

\paragraph{Dataset Challenges: }During constructing the \DatasetName{} dataset, some hard samples are observed by our annotators, as shown in Figure~\ref{fig:dataset_challenge}. For the part of hard samples, we do not exclude them from our dataset because we think that the situations represented by these samples are exactly what the model would encounter when deployed in real scenarios. Although these samples will cause difficulties for the model, if a model achieves competitive performance on our dataset, the model will ultimately not suffer much performance loss in the real application environment. Therefore, compared with previous related datasets, the fact that the data comes entirely from the real world is also a major feature and advantage of \DatasetName{}.

\section{Conclusion}
\label{sec:conclusion}
In this paper, we first pay attention to the faked character, which has never been focused on in previous Chinese Character Checking works.
To empower machines to automatically process the faked characters, we construct and annotate \DatasetName{}, a large-scale visual Chinese Character Checking dataset with faked and misspelled Chinese characters. Furthermore, we design and implement two baseline methods with different ideas. In particular, we first propose the idea of using image-text retrieval to detect the faked characters in the images.
Experimental results and detailed analyses indicate that our proposed baselines are effective and \DatasetName{} is very challenging and of great research value.
We hope our work provides better resources and a new research direction for the Chinese NLP community. In the future, we will carefully design and develop more clever models to achieve better performance on the \DatasetName{} dataset, promoting the research of writing assistance to the more realistic and intelligent stage.

\section*{Limitations}
We conduct experiments on \DatasetName{} employing two proposed baselines. Due to hardware resource limitations, we only use the base-level pre-trained weights to initialize each module in our baseline methods. In addition, because the collection and annotation of the dataset cost a lot of money, we do not have enough financial budget to test the performance of large models such as GPT-4v on our dataset. Of course, the main contribution of our work is to provide new research directions and data resources. Our designed baselines are also mainly to verify the usability of the dataset itself and to provide model design ideas for subsequent researchers to refer to. Therefore, We believe that using larger scale models to obtain better performance can be left as future work.

\section*{Ethics Statement}
In this paper, we present the human-annotated \DatasetName{}, which focuses on real-world writing assistance scenes. We have described the details of the collection, preprocessing, and annotation of our dataset in the main text of our paper. It is worth noting that all data in our dataset has obtained authorization from its providers and is desensitized before annotation to ensure that the privacy of the data providers would not be leaked. Besides, the Chinese Character Checking task itself comes from very common and important application requirements in daily life and is designed to be convenient for human daily life. Therefore, neither the task on which our work focuses nor the dataset presented poses potential harm to human society.

\end{CJK*}
\bibliography{anthology,custom}
\bibliographystyle{acl_natbib}

\appendix

\appendix


\end{document}